\documentclass[10pt,letterpaper]{article}
\usepackage[top=0.85in,left=2.75in,footskip=0.75in]{geometry}

\usepackage{amsmath,amssymb}

\usepackage{changepage}

\usepackage{textcomp,marvosym}

\usepackage{cite}

\usepackage{nameref,hyperref}

\usepackage[right]{lineno}

\usepackage[nopatch=eqnum]{microtype}
\DisableLigatures[f]{encoding = *, family = * }

\usepackage[table]{xcolor}

\usepackage{array}

\newcolumntype{+}{!{\vrule width 2pt}}

\newlength\savedwidth

\raggedright
\usepackage[aboveskip=1pt,labelfont=bf,labelsep=period,justification=raggedright,singlelinecheck=off]{caption}

\makeatletter
\renewcommand{\@biblabel}[1]{\quad#1.}
\makeatother

\usepackage{lastpage,fancyhdr,graphicx}
\usepackage{epstopdf}
\fancyheadoffset[L]{2.25in}
\fancyfootoffset[L]{2.25in}
\usepackage[utf8]{inputenc}
\usepackage[T1]{fontenc}
\usepackage{booktabs,threeparttable,tabularx,xspace,float,setspace,etoolbox}
\hypersetup{hidelinks}
\newcommand{\awarefx}{AWARE-FX\xspace}
\AtBeginEnvironment{table}{\singlespacing}
\AtBeginEnvironment{thebibliography}{\singlespacing}
\hypersetup{pdftitle={AWARE-FX: Auditable measurement of corporate foreign-exchange hedging disclosure},pdfauthor={Qi Wang},pdfsubject={Preprint}}
\begin{document}
\begin{flushleft}
{\Large\textbf{AWARE-FX: Auditable measurement of corporate foreign-exchange hedging disclosure}}
\par\bigskip
Qi Wang\textsuperscript{1*}
\par\bigskip
\textbf{1} University of Nottingham, Nottingham, United Kingdom
\par Full postal address: University of Nottingham, University Park, Nottingham NG7 2RD, United Kingdom
\par\bigskip
* Corresponding author: \href{mailto:qi.wang2@nottingham.ac.uk}{qi.wang2@nottingham.ac.uk}
\par ORCID: \href{https://orcid.org/0009-0004-8689-0733}{0009-0004-8689-0733}
\par\bigskip
Short title: Auditable measurement of hedging disclosure
\end{flushleft}
\clearpage
\newgeometry{top=0.85in,left=1in,right=1in,footskip=0.75in}
\fancyheadoffset[L]{0pt}\fancyfootoffset[L]{0pt}

\section*{Abstract}
Corporate annual reports describe hedging through transactional, policy, accounting, and explicit non-use language. This study presents \awarefx, the foreign-exchange application of the Auditable Weak-supervision and Aggregation framework, for constructing traceable firm-year disclosure measures. Professional-source retrieval, evidence-state rules, channel classifiers, and top-three aggregation link each score to supporting passages. The application covers 24,909 Hong Kong firm-years from 2008--2025 and 543,527 candidate snippets. Evaluation combines weak-label held-out tests, three-seed encoder comparisons, a 2023--2025 temporal test, calibration diagnostics, selective prediction, and a stratified 300-snippet audit coded by one researcher. The paired temporal tests do not detect statistically significant differences between FinBERT and ModernBERT. The negation-aware lexicon has higher broad-task audit F1 than strict-gate FinBERT and Qwen3-8B, demonstrating that learned models do not uniformly improve on domain rules. At 80\% temporal coverage, retained-sample F1 increases by 0.050--0.077, although the study does not measure human review time or deployment impact. Reproducible Qwen3-8B results show substantial variation across evidence channels. Linked exposure associations provide exploratory external validation, with sensitivity to the multiple-testing family. The scores measure detected hedging disclosure rather than hedge notional or causal effectiveness. The contribution is an evaluated measurement workflow with inspectable evidence and explicit validation limits. Independent duplicate annotation and retrieval-recall assessment remain necessary for stronger reliability claims.
\clearpage
\section{Introduction}

Foreign-exchange (FX) risk management is relevant to firms operating across currencies \cite{smith1985determinants,geczy1997why,bartram2009international}. Annual reports describe it through treasury policies, financial-instrument notes, derivative positions, and hedge-accounting disclosures \cite{ifrs7,ifrs9}. These disclosures differ in what they establish. A forward contract, matched-currency financing, an accounting policy, and an explicit statement of non-use cannot be treated as equivalent observations of hedging activity.

This distinction creates a text-measurement problem. Financial vocabulary is domain dependent, and the presence of a term does not establish the presence of the underlying activity \cite{loughran2011liability,loughran2016textual,chapman2001simple}. A passage may mention an unhedged exposure or explain why an economic hedge does not qualify for hedge accounting. Constructing a research variable requires a defined evidence unit, a treatment of ambiguity, and validation beyond the labels used to train a classifier \cite{grimmer2013text,gentzkow2019text}.

This paper presents \awarefx, the FX application of the Auditable Weak-supervision and Aggregation framework. It uses artificial intelligence (AI) and natural language processing (NLP) to construct traceable hedging-disclosure measures. Professional-source rules retrieve candidate passages. Status rules distinguish affirmative evidence, risk or policy context, explicit non-use, mixed statements, and accounting non-designation. Four channel classifiers rank candidates, and exact channel gates determine which probabilities enter each firm-year score. An evidence ledger retains the source report, page, passage, rule, status, model output, and aggregation contribution.

The application covers 24,909 Hong Kong firm-years from 2008--2025 and 543,527 retrieved snippets. Evaluation compares rules, sparse models, FinBERT, ModernBERT, and deterministic Qwen3-8B. It includes three-seed and forward temporal tests, probability-calibration diagnostics, selective prediction, and a stratified 300-snippet audit coded by one researcher. This audit is a diagnostic sample rather than an independently double-coded population benchmark. External exposure regressions examine associational construct validity.

The study examines three questions: how evidence-state and channel constraints change classification and score construction; how model performance varies across evidence channels and future reporting periods; and what the resulting measure establishes about disclosed hedging. Recent financial benchmarks motivate evaluation on the specific task rather than inference from general language-model capability \cite{liu2025findabench,klimaszewski2025avenibench,tang2025finmteb}.

The contribution is an evaluated, inspectable measurement workflow. It combines a documented financial evidence taxonomy, explicit aggregation rules, and model comparisons with retained source evidence. The study does not introduce a new language-model architecture or weak-supervision optimizer. Its incremental value must be assessed against simpler rules: those rules achieve higher broad-task audit F1 than the learned comparators, while channel constraints change the errors made by the encoder. Whether these outputs save analysts time remains an empirical question. The disclosure scores are intended to support text-based measurement, not estimation of hedge notional or causal hedge effectiveness.

\subsection{Related work and measurement setting}
\label{sec:related}

\subsubsection{Corporate text and hedging disclosure}

Text-as-data research treats language as a measurable object whose interpretation depends on construct definition and validation \cite{grimmer2013text,gentzkow2019text}. Financial applications measure tone, uncertainty, risk disclosure, product-market structure, and firm-level political risk \cite{tetlock2007giving,tetlock2008more,li2010textual,loughran2011liability,campbell2014information,hoberg2016text,hassan2019firm}. Domain dictionaries and contextual models address different sources of measurement error \cite{loughran2016textual,kearney2014textual}.

Hedging research studies derivatives use, financing, and exposure \cite{smith1985determinants,nance1993determinants,froot1993risk,tufano1996who,geczy1997why,allayannis2001exchange,guay2003how,bartram2009international}. Exchange-rate exposure describes the sensitivity of firm value or returns to currency movements \cite{adler1984exposure,jorion1990exchange,dominguez2006exchange,bartram2010resolving}. A disclosure measure is a different construct. It records language supporting risk-management activity and cannot establish unreported positions, derivative quantities, or causal effectiveness. Financial-reporting standards supply terminology and accounting distinctions, not gold labels for classifier evaluation \cite{ifrs7,ifrs9,ias39}.

\subsubsection{Weak labels, evidence aggregation, and language models}

Firm-year labels are coarser than the local passages that support them. Multiple-instance learning formalizes this distinction between bag labels and instance labels \cite{dietterich1997solving,maron1998framework}. Weak supervision combines imperfect labels or heuristics, while active learning directs annotation toward informative observations \cite{ratner2017snorkel,ratner2020snorkel,settles2009active}. Propagating a firm-year label to snippets can retain noise even when the train/test split keeps firm-years separate. Independent semantic annotation is therefore necessary to evaluate what the classifiers actually detect.

Contextual encoders build on transformer pretraining \cite{vaswani2017attention,devlin2019bert}. Financial-domain models and newer general-domain encoders provide alternative representations \cite{araci2019finbert,yang2020finbert,warner2024modernbert}. Generative financial benchmarks examine analytical reasoning, retrieval, multilingual embeddings, and other task-specific capabilities \cite{liu2025findabench,klimaszewski2025avenibench,tang2025finmteb}. Recent applications consider financial risk estimation, audit-risk agents, and interpretable credit-risk systems \cite{pele2026llmvar,lu2025dual,golec2026interpretable}. These studies motivate contemporary comparisons, but their scores are not directly comparable with a hedging-disclosure task using different data and labels.

Qwen3-8B supplies the primary reproducible large language model (LLM) comparator in this study \cite{yang2025qwen3}. Its model revision, deterministic decoding, prompt, and input/output hashes are archived. Results from a hosted client labelled ``Gemini 3.5 Flash High'' are retained only in \nameref{S1_Appendix}, because its exact model revision and decoding configuration could not be verified.

\subsubsection{Provenance and the evaluation boundary}

Design-science research evaluates artifacts through relevance, implementation, and evidence about their performance \cite{march1995design,hevner2004design,gregor2013positioning}. Here, provenance means that a firm-year score resolves to identifiable passages and versioned processing decisions. Negation and accounting status are explicit because a matched concept does not by itself establish affirmative evidence \cite{chapman2001simple}.

The closest comparison is between alternative ways of retrieving, classifying, and aggregating the same evidence. The present workflow contributes a documented application of these operations, not an assertion that its components are individually new. Evaluation distinguishes predictive performance, evidence traceability, and operational benefit. The first is measured on weak labels and a single-coder audit; the second is supported by the ledger; the third remains unmeasured. Claims of generalization beyond this reporting environment require separate evidence.

\section{Materials and methods}
\subsection{AWARE-FX measurement architecture}
\label{sec:architecture}

\subsubsection{Evidence records and processing rules}

\begin{figure}[H]
\centering
\includegraphics[width=\textwidth]{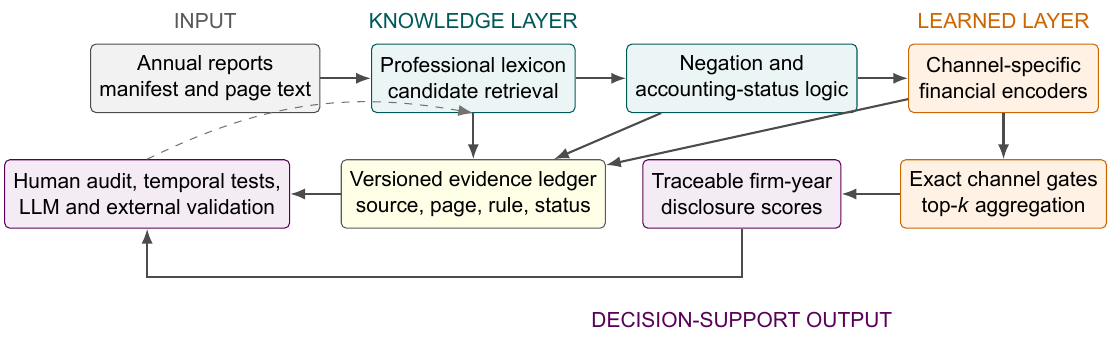}
\caption{\textbf{Measurement architecture.} Solid arrows represent data processing. The dashed audit-feedback path produces a new rule or output version; it does not convert audit labels into independent training evidence. OpenAI Codex assisted with typography and layout preparation for this submission. The author reviewed the final figure for accuracy and approved it.}
\label{fig:pipeline}
\end{figure}

\awarefx separates retrieval, status assignment, channel scoring, and firm-year aggregation. Fig~\ref{fig:pipeline} shows these operations. An input manifest identifies reports by ticker and report year. Each retrieved record retains page boundaries, passage text, matched concepts, lexicon channels, signal classes, status rules, and model probabilities. The aggregate output retains evidence counts and the passage identifiers contributing to each score.

The construction follows the artifact-evaluation perspective in design-science research \cite{march1995design,hevner2004design,gregor2013positioning}. A score must distinguish affirmative activity from policy-only, exposure-only, accounting, and non-use language. Its provenance must also identify which deterministic and learned decisions changed the result. Traceability allows inspection, although it does not by itself establish classification validity or operational efficiency.

\subsubsection{Score construction and human review}

The implemented processing rule resolves each manifest row to a report, extracts page text, consolidates nearby professional-term matches, assigns evidence status, and preserves four channel probabilities. For each specialized channel, only affirmative evidence with a compatible exact channel enters the score. The score is zero if this eligible set is empty; otherwise it is the mean of its three highest probabilities. Generic broad evidence and explicit no-use or mixed evidence counts remain separately identifiable. Section~\ref{sec:model} defines the aggregation formally.

A zero score means that the implemented pipeline found no eligible affirmative evidence. It cannot distinguish absence of disclosure from every possible retrieval, optical-character-recognition, or table-parsing failure. It also does not establish absence of actual hedging. Report-processing status and evidence counts must accompany the score when it is merged into a research panel.

Low-confidence or contradictory records can be presented for human inspection. The reported selective-prediction analysis withholds cases according to model confidence, but does not measure annotation time, the accuracy of subsequent human decisions, or a live review queue. Table~\ref{tab:system_modules} records the modules and their outputs.

\begin{table}[H]
\centering
\caption{\textbf{AWARE-FX system modules and audit artifacts}}
\label{tab:system_modules}
\begin{threeparttable}
\begin{tabularx}{\textwidth}{@{}lXX@{}}
\toprule
Module & Main function & Output artifact \\
\midrule
Manifest layer & Identifies 24,909 firm-year reports and file paths & Versioned report manifest \\
Retrieval layer & Applies an 80-rule professional-source lexicon & 543,527 candidate snippets \\
Audit layer & Tags signal class, channel, and negation/no-use status & Snippet audit fields and 300 human reviews \\
Modeling layer & Estimates four channel probabilities with selected FinBERT checkpoints & Held-out metrics and prediction audit trail \\
Aggregation layer & Separates affirmative evidence from context and no-use & Conservative firm-year disclosure panel \\
Validation layer & Tests internal classification and external screening validity & Manual metrics and construct-validation outputs \\
\bottomrule
\end{tabularx}
\begin{tablenotes}
\footnotesize
\item Note: The formal firm-year score is released only after selected-checkpoint inference with affirmative-evidence aggregation; development-stage all-candidate composites are retained for reproducibility but not used as the primary measure.
\end{tablenotes}
\end{threeparttable}
\end{table}

\subsection{Professional-source lexicon and evidence extraction}
\label{sec:lexicon}

\subsubsection{Lexicon construction}

The domain lexicon is the entry point of the formal measurement pipeline. It replaces the earlier development-stage keyword list and contains 80 versioned rules. Its terminology is grounded in IFRS 7 financial-instrument risk disclosure, IFRS 9 and IAS 39 hedge-accounting language, Big Four implementation guidance, and CFA derivatives vocabulary \cite{ifrs7,ifrs9,ias39,pwc2026hedge,ey2014hedge,kpmg2024derivatives,cfa2026currency,cfa2026swaps}. These sources do not supply empirical labels. They supply a transparent professional vocabulary for high-recall candidate retrieval.

Each rule records a canonical concept, match pattern, channel, signal class, polarity, strength, and source basis. Channels include broad hedging, FX derivatives, interest-rate derivatives, commodity derivatives, foreign-debt or natural hedging, FX-risk disclosure, and non-use. Signal classes distinguish active derivative use, hedge-accounting designation, risk-only disclosure, accounting-policy context, matched-currency financing, active natural hedging, and explicit non-use. The structure matters because the same surface term can enter different evidence states.

Foreign borrowing is included, but it is not automatically treated as hedging. A reference to a foreign-currency loan can describe funding, exposure, or a natural hedge. It becomes affirmative foreign-debt evidence only when the surrounding passage supports a matching or risk-management interpretation. Otherwise it remains a context candidate for model ranking or human review.

\subsubsection{Negation and accounting-status rules}

Fifteen status rules detect explicit no-use, discontinuation, mixed evidence, and hedge-accounting non-designation. A local cue such as ``does not use'', ``no contracts outstanding'', or ``not entered into'' generally supports no-use. In contrast, ``does not qualify for hedge accounting'' does not imply that no economic hedge exists. The rule system therefore distinguishes positive evidence with accounting non-designation from accounting-policy context and explicit non-use.

This separation prevents two asymmetric errors. Ignoring negation inflates positive disclosure by treating no-use sentences as evidence. Treating every negative grammatical construction as no-use removes genuine economic hedges that are not formally designated. Both errors are visible in annual reports and cannot be resolved by a keyword count alone.

\subsubsection{Page-level extraction}

The extraction program resolves each manifest row, parses the report page by page, and retains page boundaries. When a rule matches, the program extracts a surrounding text window and records the page number, matched text, canonical concept, channel, signal class, and source basis. Nearby matches are consolidated to reduce duplicate evidence while allowing one snippet to retain multiple concepts. The extractor writes incremental checkpoints and a progress record so that interrupted local processing can resume without rebuilding completed reports.

PDF extraction failures are recorded rather than silently dropped. The source manifest and extraction summary preserve the file status for every firm-year. Imperfect PDF syntax messages can occur in legacy reports; a report is counted as successfully processed only when usable page text and a completed summary record are produced. This distinction is relevant for coverage and survivorship assessment.

\subsubsection{Outputs and version boundary}

The two formal outputs are a snippet-level evidence table and a one-row-per-firm-year extraction summary. The evidence table is the model input and audit ledger. The summary supports corpus coverage, extraction-failure checks, and merges to firm-year data. Development-stage snippets are retained only for provenance and model development; all formal full-sample inference and reported score coverage use domain-lexicon v2 evidence extracted from the raw reports.

Across 24,909 report firm-years, the extractor processes 4,165,297 pages and returns 543,527 candidate snippets. Table~\ref{tab:lexicon_coverage} reports the status distribution. The large context-only and mixed categories are not failures of retrieval. They are deliberately retained difficult cases that permit the downstream system to learn and audit the boundary between risk discussion and affirmative hedging evidence.

\begin{table}[H]
\centering
\caption{\textbf{Domain-lexicon extraction coverage}}
\label{tab:lexicon_coverage}
\begin{threeparttable}
\footnotesize
\begin{tabularx}{\textwidth}{@{}XrrrX@{}}
\toprule
Category & Reports & Snippets & Share & Notes \\
\midrule
All processed firm-years & 24,909 & 543,527 & 100.0\% & Annual-report manifest rows processed \\
Context-only candidate & -- & 365,093 & 67.2\% & Domainlex v2 status \\
Positive candidate & -- & 157,353 & 29.0\% & Domainlex v2 status \\
Explicit no-use candidate & -- & 6,995 & 1.3\% & Domainlex v2 status \\
Mixed positive/no-use review & -- & 6,875 & 1.3\% & Domainlex v2 status \\
Positive with accounting non-designation & -- & 5,928 & 1.1\% & Domainlex v2 status \\
Accounting non-designation context & -- & 1,283 & 0.2\% & Domainlex v2 status \\
\bottomrule
\end{tabularx}
\begin{tablenotes}
\footnotesize
\item The extraction uses the professional-source domain lexicon and negation rules on raw annual reports. A snippet can be relevant for retrieval without being active hedging evidence.
\item Shares are relative to the final evidence-snippet CSV.
\end{tablenotes}
\end{threeparttable}
\end{table}

\subsection{Model training and validation}
\label{sec:model}

\subsubsection{Weak labels and model comparison}

The development data contain 76,648 weakly labeled snippets from 7,776 firm-years. Grouping by \texttt{firm\_year\_key} produces 57,232 training rows and 19,416 held-out rows. Labels originate from the earlier evidence-building stage and do not constitute gold snippet annotations. Splitting firm-years reduces within-period leakage but does not establish independence from repeated language across years of the same firm. External exposure outcomes are excluded from label construction, fitting, and checkpoint selection.

The sparse comparator is term frequency--inverse document frequency (TF--IDF) with logistic regression \cite{pedregosa2011scikit}. FinBERT initializes \texttt{BertForSequenceClassification} from \texttt{yiyanghkust/finbert-pretrain}. The archived channel run uses sequence length 256, learning rate $2\times10^{-5}$, training/evaluation batch sizes 32/64, and seed 42. Checkpoints at epochs 15, 20, 25, and 30 are compared on held-out weak labels. Selected epochs are 20 for FX and interest-rate derivatives, 25 for commodity derivatives, and 15 for foreign-debt hedging. This is checkpoint selection under fixed settings, not a comprehensive hyperparameter optimization.

Additional runs compare FinBERT and \texttt{answerdotai/ModernBERT-base} \cite{warner2024modernbert} at seeds 17, 42, and 73. The forward temporal design trains through 2021, validates on 2022, and tests on 2023--2025. The reported tests evaluate transfer across reporting years within the same market; they do not demonstrate transfer to other jurisdictions or languages. Runtime and memory measurements are archived, but no end-to-end deployment cost or matched hyperparameter-search budget is established in the present manuscript.

\subsubsection{Statistical comparison and selective prediction}

The encoder tables use two estimands. The multi-seed table reports the mean of the three separately computed F1 scores. The paired temporal table first averages the three predicted probabilities within each encoder family, applies threshold 0.5, and then computes ensemble F1. Averaging F1 and computing F1 after averaging probabilities need not give the same value. The FX values 0.872 and 0.880 refer to these respective procedures.

Paired encoder probabilities are aligned on the same 13,544 temporal observations. F1 differences use 2,000 paired bootstrap resamples, while exact two-sided McNemar tests compare discordant correctness outcomes. These archived intervals resample snippets, not firm or firm-year clusters; dependence within those groups may affect uncertainty. The absence of a detected difference is not an equivalence test. Area under the receiver operating characteristic curve (ROC--AUC), Brier score, and ten-bin expected calibration error (ECE) describe ranking and calibration under the available labels.

Selective prediction orders observations by $|p_i-0.5|$ and withholds the least-confident 5\%, 10\%, 20\%, or 30\%. Retained-sample performance is reported with coverage. These results describe the model's retained cases, not the accuracy or cost of a completed human-review workflow.

\subsubsection{Channel gates and firm-year aggregation}

Let $p_{ijc}$ be the probability for snippet $j$ of firm-year $i$ and channel $c$, and $E_{ic}$ the set of affirmative snippets with an exact compatible channel. Let $K_{ic}$ contain the top three probabilities in $E_{ic}$, or the entire set if it has fewer than three members. The strict disclosure score is
\begin{equation}
S_{ic}=\begin{cases}
0, & |E_{ic}|=0,\\
|K_{ic}|^{-1}\displaystyle\sum_{j\in K_{ic}}p_{ijc}, & |E_{ic}|>0.
\end{cases}
\end{equation}
Generic broad candidates are excluded from the four specialized channel sets and retained for separate broad measures. Context-only, explicit no-use, mixed, and accounting-non-designation-context records do not enter affirmative scores. Affirmative economic-hedge evidence without accounting designation remains eligible.

The top-three choice is an implemented measurement rule. The current evidence does not establish that three is optimal or that this rule is robust across values of $k$. Repeated text and retrieval coverage can affect a firm's opportunity to obtain a high score. Sensitivity to aggregation and omitted evidence is a remaining validation requirement.

\subsubsection{Human audit and reproducible LLM evaluation}

The 300-snippet audit oversamples informative candidate statuses. One researcher assigned the final channel, broad, non-use, context, uncertainty, and accounting labels with AI-assisted consistency review. It is not independent duplicate annotation and provides no inter-rater reliability statistic. Audit precision and accuracy are conditional on this sampling design, and neither weak-label tests nor this enriched audit estimate population accuracy. Reusing these observations for system refinement also limits their status as a final untouched benchmark.

Qwen3-8B is the primary reproducible generative comparator \cite{yang2025qwen3}. It receives the fixed blinded input and label schema, with deterministic non-sampling decoding and thinking disabled. The archived run records model revision, package versions, input and prompt hashes, GPU, runtime, and memory. All 300 IDs, hashes, and output schemas pass the archived consistency checks. Its labels are compared with the same single-coder audit, so reproducibility does not remove the audit's validation limits. The hosted comparator is described separately in \nameref{S1_Appendix}.

Table~\ref{tab:model_benchmark} presents the archived results.

\begin{table}[H]
\centering
\caption{\textbf{Selected held-out FinBERT checkpoints after platform-epoch assessment}}
\label{tab:model_benchmark}
\begin{threeparttable}
\footnotesize
\begin{tabularx}{\textwidth}{@{}XrrrrX@{}}
\toprule
Channel & Precision & Recall & F1 & AUC & Status \\
\midrule
Commodity derivatives (epoch 25) & 0.875 & 0.766 & 0.817 & 0.933 & selected \\
FX derivatives (epoch 20) & 0.912 & 0.893 & 0.902 & 0.945 & selected \\
Foreign debt hedging (epoch 15) & 0.829 & 0.767 & 0.797 & 0.912 & selected \\
Interest rate derivatives (epoch 20) & 0.913 & 0.884 & 0.898 & 0.950 & selected \\
\bottomrule
\end{tabularx}
\begin{tablenotes}
\footnotesize
\item Metrics are held-out snippet-classification results. The platform run evaluated epochs 15, 20, 25, and 30; its mean F1 was 0.852, 0.852, 0.851, and 0.851, respectively.
\item Checkpoint choice uses held-out performance only. The selected combination is separately checked against the 300 human-audited snippets before the formal score release.
\end{tablenotes}
\end{threeparttable}
\end{table}

Table~\ref{tab:manual_validation} presents the archived results.

\begin{table}[H]
\centering
\caption{\textbf{Manual audit of domain-lexicon v2 snippets}}
\label{tab:manual_validation}
\begin{threeparttable}
\footnotesize
\begin{tabularx}{\textwidth}{@{}Xrrrrrr@{}}
\toprule
Candidate status & Audit $N$ & True evidence & True rate & No-use & Context-only & \multicolumn{1}{c@{}}{} \\
\midrule
Positive candidate & 140 & 108 & 77.1\% & 15 & 20 \\
Context-only candidate & 70 & 7 & 10.0\% & 13 & 57 \\
Explicit no-use candidate & 35 & 1 & 2.9\% & 34 & 34 \\
Mixed positive/no-use review & 30 & 3 & 10.0\% & 24 & 24 \\
Positive with accounting non-designation & 20 & 15 & 75.0\% & 0 & 0 \\
Accounting non-designation context & 5 & 5 & 100.0\% & 0 & 0 \\
\addlinespace
\multicolumn{7}{@{}l}{\textit{Panel B: selected checkpoints with strict channel gates, threshold = 0.5}} \\
Channel & Human + & Pred. + & Precision & Recall & F1 & AUC \\
FX derivatives & 54 & 33 & 0.818 & 0.500 & 0.621 & 0.738 \\
Interest-rate derivatives & 19 & 27 & 0.556 & 0.789 & 0.652 & 0.877 \\
Commodity derivatives & 21 & 19 & 0.789 & 0.714 & 0.750 & 0.874 \\
Foreign-debt hedging & 25 & 23 & 0.870 & 0.800 & 0.833 & 0.934 \\
Broad-any, strict or generic evidence & 139 & 141 & 0.780 & 0.791 & 0.786 & 0.838 \\
\bottomrule
\end{tabularx}
\begin{tablenotes}
\footnotesize
\item The table reports a stratified human audit of 300 domain-lexicon v2 snippets. Rates are audit confirmation rates within the sampled strata, not population prevalence estimates.
\item Panel B evaluates the channel-specific selected checkpoints after exact lexicon-channel gating. Generic \texttt{broad\_hedging} candidates are reported through the broad-any measure rather than admitted to every specialised channel.
\item Final labels were manually reviewed by the author and then screened through an AI-assisted consistency check; final labels are human-adjudicated.
\end{tablenotes}
\end{threeparttable}
\end{table}

Table~\ref{tab:baseline_ablation} presents the archived results.

\begin{table}[H]
\centering
\caption{\textbf{Baseline and ablation comparison on the 300-snippet human audit}}
\label{tab:baseline_ablation}
\begin{threeparttable}
\footnotesize
\begin{tabularx}{\textwidth}{@{}Xrrrr@{}}
\toprule
\multicolumn{5}{@{}l}{\textit{Panel A: broad hedging evidence}} \\
Method & Precision & Recall & F1 & AUC \\
\midrule
Keyword pattern & 0.513 & 0.978 & 0.673 & 0.589 \\
Domain lexicon without negation & 0.717 & 0.892 & 0.795 & 0.794 \\
Domain lexicon with negation & 0.766 & 0.871 & \textbf{0.815} & 0.820 \\
TF--IDF logistic regression & 0.684 & 0.921 & 0.785 & \textbf{0.868} \\
FinBERT without strict gate & 0.631 & 0.899 & 0.742 & 0.828 \\
FinBERT with strict gate & 0.780 & 0.791 & 0.786 & 0.838 \\
Qwen3-8B & -- & -- & 0.803 & -- \\
\addlinespace
\multicolumn{5}{@{}l}{\textit{Panel B: channel-specific F1}} \\
Method & FX & Interest rate & Commodity & Foreign debt \\
\midrule
Keyword pattern & 0.484 & 0.717 & 0.857 & 0.840 \\
Domain lexicon without negation & 0.615 & 0.652 & 0.780 & \textbf{0.880} \\
Domain lexicon with negation & 0.607 & 0.652 & 0.780 & \textbf{0.880} \\
TF--IDF logistic regression & 0.663 & 0.362 & 0.526 & 0.430 \\
FinBERT without strict gate & 0.568 & 0.319 & 0.606 & 0.532 \\
FinBERT with strict gate & 0.621 & 0.652 & 0.750 & 0.833 \\
Qwen3-8B & \textbf{0.791} & \textbf{0.826} & \textbf{0.952} & 0.148 \\
\bottomrule
\end{tabularx}
\begin{tablenotes}
\footnotesize
\item All methods use the same single-coder audit labels. Probabilistic classifiers use threshold 0.5; rules and Qwen3-8B supply hard labels. The stratified audit does not estimate population predictive values. Unreported Qwen precision, recall, and AUC are not imputed.
\item TF--IDF models use 76,645 weakly labelled training snippets after removing three exact text matches with the audit sample. The comparison evaluates complementary system components rather than presuming transformer dominance.
\item Qwen3-8B uses the archived deterministic run. Hosted-client comparisons are supplementary exploratory evidence in \nameref{S1_Appendix}.
\end{tablenotes}
\end{threeparttable}
\end{table}

\subsection{External construct-validation design}
\label{sec:external}

The exposure analysis examines associations between detected disclosure and FX exposure, following the exposure literature \cite{adler1984exposure,jorion1990exchange,dominguez2006exchange,bartram2010resolving}. It does not estimate causal hedging effects. Reporting incentives, risk levels, financing, and unobserved treasury practices can jointly affect disclosure and exposure.

Exposure estimates vary by firm, year, and currency label. Of 24,909 report firm-years, 24,435 merge to 51,140 exposure rows. Complete-data all-successful regressions use 49,970 rows; the main sample contains 21,325 firm-years and 44,582 rows. Multiple currency labels can represent the same firm-year. Table~\ref{tab:sample_flow} reports this distinction.

\begin{table}[H]
\centering
\caption{\textbf{Flow from annual-report firm-years to exposure-validation records}}
\label{tab:sample_flow}
\begin{threeparttable}
\footnotesize
\begin{tabularx}{\textwidth}{@{}Xrr@{}}
\toprule
Stage & Firm-years & Firm-year--FX-label rows \\
\midrule
Domain-lexicon v2 score release & 24,909 & 24,909 \\
Merged to exposure panel & 24,435 & 51,140 \\
All-successful complete regression sample & -- & 49,970 \\
Predefined main regression sample after merge & 21,325 & 44,582 \\
\bottomrule
\end{tabularx}
\begin{tablenotes}
\footnotesize
\item One report firm-year can map to multiple successful currency or index exposure estimates. The disclosure score and controls vary by firm-year; outcomes also vary by FX label. Standard errors are clustered by ticker.
\item A firm-year count is not reported for the complete-case subset because the archived regression output records the complete row count; 24,435 and 21,325 are verified merge-stage firm-year counts.
\end{tablenotes}
\end{threeparttable}
\end{table}

For firm $i$, year $t$, and FX label $l$, the archived model is
\begin{equation}
Y_{itl}=\alpha+\beta S^{FX}_{it}+\mathbf{X}_{it}'\gamma+\lambda_l+\tau_t+\varepsilon_{itl},
\end{equation}
where $Y_{itl}$ is baseline, stress-period, or incremental absolute exposure. Controls are size, leverage, quick ratio, and book-to-market ratio. FX-label and year effects enter the model, and standard errors are clustered by ticker. The broad-any score is evaluated in otherwise corresponding regressions. Firm clustering addresses dependence in standard errors; it does not absorb time-invariant firm heterogeneity. A firm-fixed-effects sensitivity analysis has not yet been completed for this revision.

All tests are exploratory. Bonferroni correction does not require preregistration, but the choice of a test family must be transparent and the adjustment does not make retrospectively selected analyses confirmatory. At family size six for strict-FX tests, the threshold is $0.05/6=0.00833$. The two stress estimates and main-sample baseline estimate pass this threshold; the all-successful baseline estimate ($p=0.0089$) and both incremental estimates do not. Including the six broad-score tests gives $0.05/12=0.00417$, passed only by the two strict-FX stress estimates. These checks do not cover all model choices explored during development.

One score yielding a smaller $p$-value than another does not establish a statistically significant difference between their coefficients. The reported comparison provides a descriptive specificity check, subject to panel dependence, sample selection, and omitted variables. Table~\ref{tab:external_validation} reports the unchanged archived estimates.

\begin{table}[H]
\centering
\caption{\textbf{External construct validation using FX exposure outcomes}}
\label{tab:external_validation}
\begin{threeparttable}
\footnotesize
\begin{tabularx}{\textwidth}{@{}lXrrrr@{}}
\toprule
Sample & Outcome & Coefficient & Clustered SE & $p$-value & $N$ \\
\midrule
\multicolumn{6}{@{}l}{\textit{Panel A: strict FX disclosure score}} \\
All successful & Baseline absolute exposure & -0.236 & 0.090 & 0.0089 & 49,970 \\
All successful & Stress-period absolute exposure & -0.374 & 0.111 & 0.0008 & 49,970 \\
All successful & Incremental stress exposure & -0.466 & 0.199 & 0.0195 & 49,970 \\
Main sample & Baseline absolute exposure & -0.254 & 0.090 & 0.0046 & 44,582 \\
Main sample & Stress-period absolute exposure & -0.427 & 0.110 & 0.0001 & 44,582 \\
Main sample & Incremental stress exposure & -0.408 & 0.208 & 0.0501 & 44,582 \\
\addlinespace
\multicolumn{6}{@{}l}{\textit{Panel B: broad-any disclosure score}} \\
All successful & Baseline absolute exposure & -0.046 & 0.086 & 0.5880 & 49,970 \\
All successful & Stress-period absolute exposure & -0.048 & 0.112 & 0.6679 & 49,970 \\
All successful & Incremental stress exposure & -0.356 & 0.207 & 0.0856 & 49,970 \\
Main sample & Baseline absolute exposure & -0.085 & 0.083 & 0.3062 & 44,582 \\
Main sample & Stress-period absolute exposure & -0.115 & 0.109 & 0.2927 & 44,582 \\
Main sample & Incremental stress exposure & -0.247 & 0.212 & 0.2442 & 44,582 \\
\bottomrule
\end{tabularx}
\begin{tablenotes}
\footnotesize
\item Each panel uses the named continuous score. Models include size, leverage, quick ratio, book-to-market ratio, FX-label effects, and year effects; standard errors are clustered by ticker.
\item These are exploratory associations. With six Panel-A tests, the threshold is 0.00833: both stress estimates and the main-sample baseline estimate pass, while the all-successful baseline ($p=0.0089$) and both incremental estimates do not. With twelve tests across both panels, only the two strict-FX stress estimates pass 0.00417. Firm effects are absent; raw coefficients are unchanged. Adjustment does not make the analyses preregistered or confirmatory.
\end{tablenotes}
\end{threeparttable}
\end{table}

\subsection{Use of artificial intelligence tools}
FinBERT, ModernBERT and Qwen3-8B are evaluated research models. The hosted Gemini comparator is exploratory, with the identification and configuration limits stated in \nameref{S1_Appendix}. Model outputs are assessed against the archived evaluation labels; these labels and comparisons retain the validation limits described above.

The author manually labelled the audit sample and retained responsibility for the final labels. A subsequent consistency review used OpenAI Codex (recorded session model identifier: \texttt{gpt-5.5}; July 2026) with local Python scripts to check completeness, permissible label values, cross-label conflicts, and agreement between notes and labels. Flagged cases were examined and corrections were made with the author's approval. The precise tool and model used for an earlier AI look-over of the author-labelled workbook were not established from the available archive. These checks do not constitute independent duplicate human annotation or establish inter-rater reliability.

During software and manuscript preparation, OpenAI Codex was used for code writing and review, language editing, LaTeX formatting, figure rendering and layout, bibliographic metadata checking, and preparation of the data and code archive. Google Gemini via Antigravity was used for language editing, LaTeX formatting, code review and plotting assistance. Anthropic Claude was used to check the submission materials for consistency, to draft the documentation of the data and code archive, and to draft the wording of this paragraph and of one paragraph in the Limitations subsection. Exact tool versions were not recorded except where stated. The author reviewed all AI-assisted content, verified reported numbers against archived outputs and references against their sources, and takes full responsibility for the work.
\section{Results}
\label{sec:results}

\subsection{Retrieval and score construction}

The release reports 543,527 snippets from 24,909 firm-years and 4,165,297 parsed pages. The six status counts in Table~\ref{tab:lexicon_coverage} sum to the snippet total: 365,093 context-only, 157,353 positive, 6,995 explicit non-use, 6,875 mixed, 5,928 positive with accounting non-designation, and 1,283 accounting-non-designation-context records. These are mutually exclusive candidate-status rows in the reported table. Earlier term-, rule-, and channel-event counters use different units and are not substituted for snippet counts. The status reconciliation checks the reported table; a complete raw-output counter audit remains required.

Selected-checkpoint inference produces 24,909 firm-year score rows. During development, admitting generic broad candidates to all specialized channels produced 74--76\% candidate coverage and score saturation. Exact channel gating corrects that construction: nonzero shares become 22.8\% for FX, 18.2\% for interest-rate, 2.8\% for commodity, and 2.5\% for foreign-debt scores. Shares at or above 0.5 are 18.0\%, 18.1\%, 2.6\%, and 2.2\%. These are score distributions, not validated estimates of actual hedging prevalence.

\subsection{Encoder comparison and temporal reliability}

The archived epoch comparison shows a mean held-out F1 increase of 0.000183 between epochs 15 and 20, below the implemented 0.002 selection threshold. Channel-selected weak-label F1 is 0.902 for FX, 0.898 for interest rate, 0.817 for commodity, and 0.797 for foreign debt. These values concern weak development labels, distinct from the semantic audit results.

Table~\ref{tab:encoder_multiseed} reports mean F1 across three separately trained seeds. Grouped-random FinBERT values are 0.894, 0.891, 0.768, and 0.793 across the four channels; the corresponding ModernBERT values are 0.894, 0.884, 0.761, and 0.780. Forward testing lowers FinBERT F1 to 0.872, 0.835, 0.711, and 0.702. Foreign-debt performance declines most under the temporal design.

\begin{table}[H]
\centering
\caption{\textbf{Multi-seed encoder performance under grouped-random and temporal evaluation}}
\label{tab:encoder_multiseed}
\begin{threeparttable}
\scriptsize
\setlength{\tabcolsep}{4pt}
\begin{tabularx}{\textwidth}{@{}llrrrr@{}}
\toprule
Task & Split & FinBERT F1 & ModernBERT F1 & Difference & Higher mean \\
\midrule
FX & Grouped random & 0.894 & 0.894 & $-0.000$ & FinBERT \\
FX & Temporal & 0.872 & 0.870 & $-0.002$ & FinBERT \\
Interest rate & Grouped random & 0.891 & 0.884 & $-0.007$ & FinBERT \\
Interest rate & Temporal & 0.835 & 0.836 & 0.001 & ModernBERT \\
Commodity & Grouped random & 0.768 & 0.761 & $-0.007$ & FinBERT \\
Commodity & Temporal & 0.711 & 0.698 & $-0.014$ & FinBERT \\
Foreign debt & Grouped random & 0.793 & 0.780 & $-0.013$ & FinBERT \\
Foreign debt & Temporal & 0.702 & 0.699 & $-0.003$ & FinBERT \\
\bottomrule
\end{tabularx}
\begin{tablenotes}
\scriptsize
\item Values are means of separately computed F1 across seeds 17, 42, and 73. Difference is ModernBERT minus FinBERT. Higher mean is descriptive and may depend on unrounded values; it does not establish significance or a preferred model. The temporal design trains through 2021, validates on 2022, and tests on 2023--2025 reports. Probability-ensemble F1 is reported separately in Table~\ref{tab:paired_temporal}.
\end{tablenotes}
\end{threeparttable}
\end{table}

Table~\ref{tab:paired_temporal} evaluates three-seed probability ensembles. Its FX F1 of 0.880 is an ensemble statistic, unlike the mean-per-seed value 0.872 above. ModernBERT-minus-FinBERT differences range from $-0.011$ to 0.008. All archived paired bootstrap intervals include zero; exact McNemar $p$-values range from 0.099 to 0.652. No statistically significant difference is detected by these tests. This result neither establishes FinBERT superiority nor demonstrates formal equivalence, and the snippet-resampling intervals do not account for all possible firm dependence.

\begin{table}[H]
\centering
\caption{\textbf{Paired three-seed probability-ensemble comparison on the temporal test}}
\label{tab:paired_temporal}
\begin{threeparttable}
\scriptsize
\setlength{\tabcolsep}{3pt}
\begin{tabularx}{\textwidth}{@{}Xrrrrr@{}}
\toprule
Task & FinBERT F1 & ModernBERT F1 & Difference & 95\% paired CI & McNemar $p$ \\
\midrule
FX derivatives & 0.880 & 0.878 & $-0.002$ & [$-0.006$, 0.002] & 0.182 \\
Interest-rate derivatives & 0.844 & 0.843 & $-0.001$ & [$-0.006$, 0.003] & 0.652 \\
Commodity derivatives & 0.731 & 0.720 & $-0.011$ & [$-0.022$, 0.001] & 0.099 \\
Foreign-debt hedging & 0.715 & 0.723 & 0.008 & [$-0.001$, 0.017] & 0.255 \\
\bottomrule
\end{tabularx}
\begin{tablenotes}
\scriptsize
\item Probabilities are averaged across seeds 17, 42, and 73 within each family before thresholding at 0.5 and calculating F1. This differs from averaging the three per-seed F1 scores. Difference is ModernBERT minus FinBERT. Intervals use 2,000 paired snippet resamples of the same 13,544 temporal observations, without firm-cluster resampling. McNemar tests are exact and two-sided; no difference detected is not proof of equivalence.
\end{tablenotes}
\end{threeparttable}
\end{table}

\begin{figure}[H]
\centering
\includegraphics[width=\textwidth]{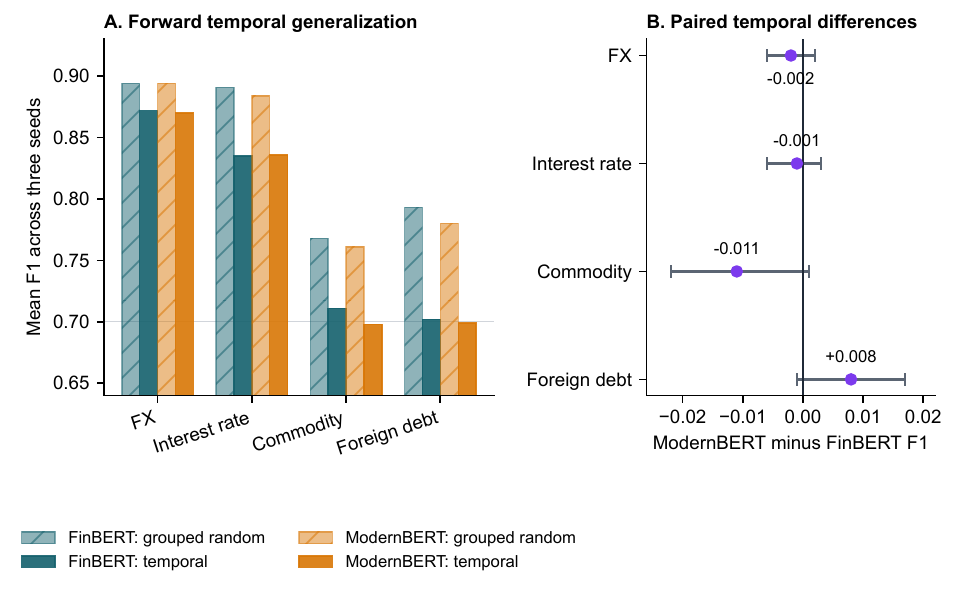}
\caption{\textbf{Encoder evaluation.} Panel A reports mean per-seed F1. Panel B reports paired differences between three-seed probability ensembles. The two panels use different aggregation estimands. OpenAI Codex assisted with typography and layout preparation for this submission. The author reviewed the final figure for accuracy and approved it.}
\label{fig:encoder_generalization}
\end{figure}

Fig~\ref{fig:encoder_generalization} illustrates the encoder comparisons.

Temporal ECE increases from 0.093 to 0.137 for FX and from 0.100 to 0.152 for foreign debt. At 80\% retained coverage, F1 increases by 0.050, 0.058, 0.077, and 0.064 across the four channels. This gain is conditional on excluding low-confidence cases. It is not an estimate of complete-workflow accuracy or analyst time saved.

\begin{table}[H]
\centering
\caption{\textbf{Temporal generalization and selective prediction for FinBERT}}
\label{tab:temporal_selective}
\begin{threeparttable}
\footnotesize
\begin{tabularx}{\textwidth}{@{}Xrrrr@{}}
\toprule
Task & Random F1 & Temporal F1 & Relative change & F1 at 80\% coverage \\
\midrule
FX derivatives & 0.894 & 0.872 & $-2.5\%$ & 0.923 \\
Interest-rate derivatives & 0.891 & 0.835 & $-6.2\%$ & 0.893 \\
Commodity derivatives & 0.768 & 0.711 & $-7.5\%$ & 0.788 \\
Foreign-debt hedging & 0.793 & 0.702 & $-11.5\%$ & 0.765 \\
\bottomrule
\end{tabularx}
\begin{tablenotes}
\footnotesize
\item Selective prediction abstains on the 20\% least-confident temporal-test observations. Performance at reduced coverage describes retained observations and is not a full-coverage population metric.
\end{tablenotes}
\end{threeparttable}
\end{table}

\begin{figure}[H]
\centering
\includegraphics[width=\textwidth]{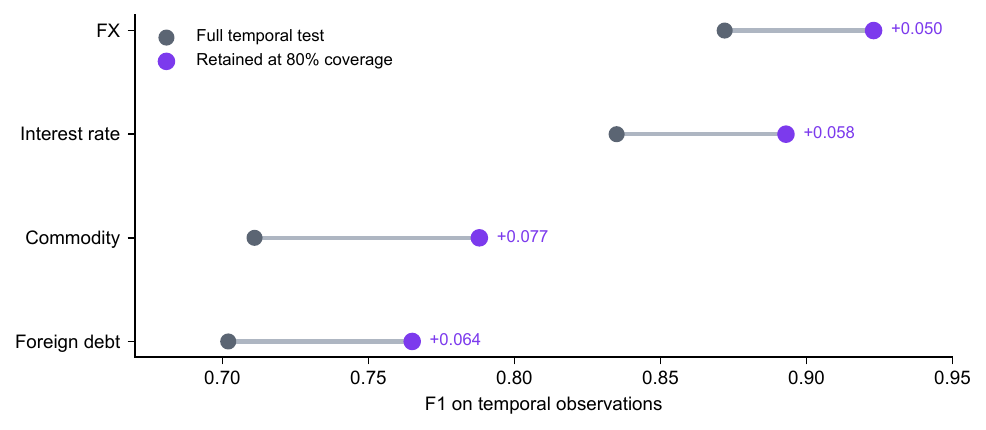}
\caption{\textbf{Selective prediction on the temporal test.} F1 at 80\% coverage describes retained cases; the excluded 20\% still require a separate review or abstention policy. Gain annotations use the archived unrounded F1 differences. OpenAI Codex assisted with typography and layout preparation for this submission. The author reviewed the final figure for accuracy and approved it.}
\label{fig:selective_prediction}
\end{figure}

Fig~\ref{fig:selective_prediction} illustrates the archived selective-prediction results.

\subsection{Human audit, rules, and Qwen3-8B}

The enriched 300-row audit contains 139 broad-positive, 135 context-only, 86 no-use or negation, and 24 unclear labels. These semantic flags overlap and do not form a mutually exclusive status partition. The results remain conditional on one researcher's labels and the sampling design.

The negation-aware lexicon reaches broad F1 of 0.815, compared with 0.795 without negation, 0.673 for keywords, 0.785 for TF--IDF, and 0.786 for strict-gate FinBERT. Thus, the learned models do not beat the rule comparator on this broad-label metric. FinBERT's strict gate raises precision from 0.631 to 0.780 and F1 from 0.742 to 0.786 relative to its ungated version, isolating a within-model effect of the implemented semantic constraint. It does not establish that the full hybrid system outperforms every rule baseline. Channel-level audit F1 also varies: strict FinBERT obtains 0.621 for FX, 0.652 for interest rate, 0.750 for commodity, and 0.833 for foreign debt.

Table~\ref{tab:llm_audit} reports the primary generative comparator. Deterministic Qwen3-8B obtains broad F1 of 0.803 and high commodity and negation/non-use F1 of 0.952 and 0.911, but foreign-debt F1 is 0.148. Context-only and accounting-only F1 are 0.548 and 0.408. These task differences favor reporting separate evidence categories rather than a single LLM capability claim. Rule sensitivity, weak-label noise, retrieval gates, and enriched audit selection are plausible explanations for the ranking differences; independent labels are needed to distinguish them.

\begin{table}[H]
\centering
\caption{\textbf{Reproducible Qwen3-8B comparison on the stratified single-coder audit}}
\label{tab:llm_audit}
\begin{threeparttable}
\footnotesize
\begin{tabularx}{\textwidth}{@{}Xr@{}}
\toprule
Audit label & Qwen3-8B F1 \\
\midrule
FX derivatives & 0.791 \\
Interest-rate derivatives & 0.826 \\
Commodity derivatives & 0.952 \\
Foreign-debt hedging & 0.148 \\
Broad hedging & 0.803 \\
Negation/non-use & 0.911 \\
Context-only disclosure & 0.548 \\
Hedge-accounting only & 0.408 \\
Unclear & 0.108 \\
\bottomrule
\end{tabularx}
\begin{tablenotes}
\footnotesize
\item Archived deterministic decoding and a recorded model revision support reproducibility. Stratified sampling and single-coder labels limit validity and population interpretation. Hosted-client results are in \nameref{S1_Appendix}.
\end{tablenotes}
\end{threeparttable}
\end{table}

Table~\ref{tab:error_modes} summarizes observed error categories and the processing response. It is a qualitative diagnostic, not an estimate of their population frequencies.
\begin{table}[H]
\centering
\caption{\textbf{Observed evidence-classification failure modes and system responses}}
\label{tab:error_modes}
\begin{threeparttable}
\footnotesize
\begin{tabularx}{\textwidth}{@{}p{3.0cm}XX@{}}
\toprule
Failure mode & Consequence & AWARE-FX response \\
\midrule
Generic risk or policy language & Keyword hit is treated as executed hedging & Context-only status, human audit, and probability ranking \\
Explicit non-use containing hedge terms & Positive score is assigned to a firm that states no use & Local no-use rules and exclusion from affirmative aggregation \\
Economic hedge without accounting designation & Genuine activity is removed by a simple negation rule & Separate positive-with-non-designation status \\
Generic broad evidence propagated to specialised channels & Cross-channel coverage and score saturation & Exact channel gates and a separate broad-any score \\
Repeated boilerplate within a report & One policy paragraph dominates the firm-year measure & Consolidated snippets and top-three mean aggregation \\
Hosted LLM overconfidence & Confidence cannot rank uncertain review cases & Treat hard labels as a benchmark; do not use hosted confidence as calibrated probability \\
Rare or semantically broad foreign-debt evidence & Model families disagree and recall can collapse & Preserve source passages, report channel metrics, and route disagreements for review \\
\bottomrule
\end{tabularx}
\begin{tablenotes}
\footnotesize
\item Failure modes are derived from the aggregation audit, the 300-snippet human review, and the fixed-prompt hosted-model benchmark.
\end{tablenotes}
\end{threeparttable}
\end{table}

\subsection{Exploratory exposure associations}

The main-sample strict-FX coefficients are $-0.254$ for baseline absolute exposure ($p=0.0046$), $-0.427$ for stress-period exposure ($p<0.001$), and $-0.408$ for incremental exposure ($p=0.0501$). With six strict-FX tests, the main-sample baseline and both samples' stress estimates pass the Bonferroni threshold. The all-successful baseline estimate ($p=0.0089$) does not. Under the twelve-test family including broad-score tests, only the two strict-FX stress associations pass.

\begin{figure}[H]
\centering
\includegraphics[width=\textwidth]{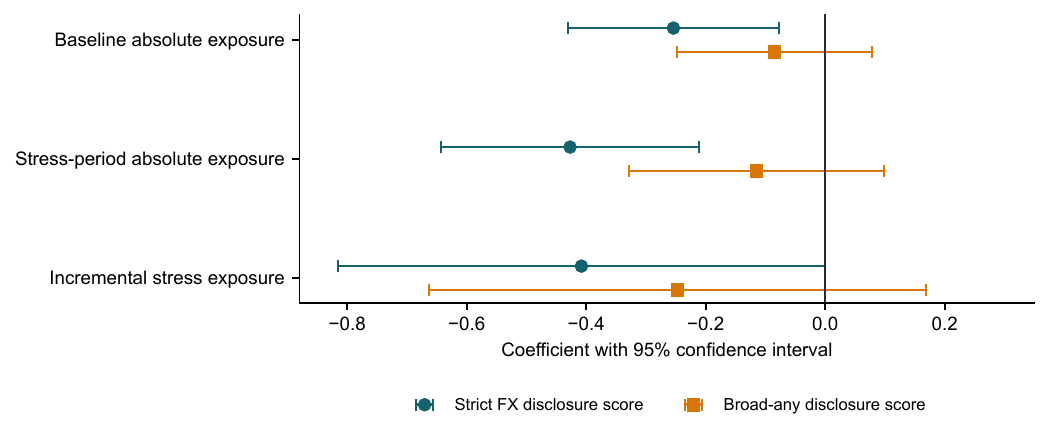}
\caption{\textbf{Archived external-validation estimates in the main sample with ticker-clustered 95\% intervals.} The regressions include FX-label and year effects but not firm effects. Coefficient differences between score specifications have not been formally tested. OpenAI Codex assisted with typography and layout preparation for this submission. The author reviewed the final figure for accuracy and approved it.}
\label{fig:external_construct_validation}
\end{figure}

Broad-any scores do not have statistically significant baseline or stress associations in the archived models. Different significance outcomes across the two score regressions do not establish a difference between coefficients. Fig~\ref{fig:external_construct_validation} displays estimates and uncertainty. Firm-fixed-effects and alternative aggregation checks remain necessary before stronger construct-validity claims. All reported coefficients describe associations.

\section{Discussion and conclusion}
\label{sec:discussion}

\subsection{Measurement value and model limits}

The available evidence supports explicit treatment of financial language and evidence status. Negation improves the rule baseline, and exact channel gates improve the encoder's audit precision. A source ledger exposes how these decisions enter the score. This is an application-level measurement contribution, not evidence of a new learning algorithm or universal classifier advantage. Recent financial benchmarks similarly distinguish task-specific performance from general language-model capability \cite{liu2025findabench,tang2025finmteb,golec2026interpretable}.

The negation-aware lexicon's broad F1 exceeds that of strict FinBERT and Qwen3-8B on the audit. The neural component is therefore optional for applications that require only this broad classification on similar evidence. The present results do not establish that continuous model scores or auditability justify additional cost in every setting. This justification would require independent score-level validation and measured reviewer workload at a specified quality target.

The paired temporal tests detect no significant encoder difference, while both encoders degrade on future-period weak labels. Selective prediction improves retained-case metrics but leaves the excluded cases unresolved. Exposure associations provide an additional descriptive check, with sensitivity to the test family and omitted firm effects. Their interpretation remains separate from causal hedging effectiveness.

\subsection{Limitations and validation requirements}

The principal semantic-validation limitation is the single-coder, enriched 300-snippet audit. AI-assisted consistency review does not supply a second independent human judgment. Inter-rater reliability, disagreement adjudication, and a further independently sampled final test are needed. Because the audit informs development, freezing models and gates before a new evaluation is important. Weak-label test performance cannot substitute for those labels, and the archived paired bootstrap resamples do not account for every source of firm dependence.

Retrieval is another unvalidated stage. Auditing retrieved snippets alone cannot estimate evidence missed outside the candidate set. Scanned reports, extraction failures, differently structured notes, and derivative tables may contain relevant evidence that the text pipeline omits. Missing table structure can lose notional values, fair values, and links between instrument type and purpose. This is a present coverage limitation. Zero-if-no-evidence scores must be interpreted alongside extraction status and cannot establish non-use.

The top-three aggregation parameter has not yet passed a sensitivity analysis across alternative $k$ values. Differences in report length, duplicated boilerplate, and snippet eligibility may affect firm-year scores. The external regressions omit firm effects and are exploratory. Clustering standard errors by ticker does not resolve confounding or selection into successful first-stage exposure estimates.

The archived scoring pipeline has three further implementation limits. Each snippet enters channel aggregation only through its primary channel, defined by the first lexicon match; 57,218 of the 543,527 snippets match more than one channel, and their evidence for the other channels is not credited. Candidate status is derived from lexicon matches grouped within 120 characters of the primary match, whereas the snippet passed to the encoder spans up to 360 characters on either side of it, so matches elsewhere in the snippet may not contribute to its status. One firm-year whose report file is missing receives zero rather than missing scores. A revised aggregation addressing these points was developed after the reported analyses were frozen; it reuses the same model probabilities, has not been independently validated, and is not used here.

The corpus is restricted to the studied English-language Hong Kong reporting environment. Temporal testing provides within-market evidence; multilingual or cross-market performance remains unknown. Rule maintenance, changes in accounting practice, and novel instruments can alter retrieval behavior. End-to-end inference cost, analyst review time, and operational effects are not established. The hosted generative result has incomplete model-identification metadata and is confined to exploratory supplementary evidence.

\subsection{Conclusion}

\awarefx constructs traceable firm-year hedging-disclosure scores from professional retrieval rules, explicit evidence states, channel classification, and aggregation. The application demonstrates inspectable score construction and heterogeneous performance across channels and time. Rule baselines remain competitive, and the paired encoder tests do not establish superiority. Independent annotation, retrieval-recall assessment, aggregation sensitivity, and operational evaluation define the remaining work needed to substantiate stronger reliability and usefulness claims. The score measures detected disclosure rather than unobserved hedging positions or their causal effectiveness.

\section*{Acknowledgments}
None.

\section*{Supporting information}
\paragraph*{S1 Appendix.}
\label{S1_Appendix}
\textbf{Exploratory hosted comparator and archived training details.}
The hosted-client comparison and archived epoch details are given below. The hosted model revision and decoding configuration remain unverified.
\clearpage
\renewcommand{\thetable}{A}
\section*{S1 Appendix. Exploratory hosted comparator and archived training details}
\label{sec:hosted_exploratory}

The hosted client identified its model as ``Gemini 3.5 Flash High'', but did not expose a stable API model identifier, temperature, seed, or complete system configuration. This displayed name is not a verified model revision. Frozen inputs, prompts, outputs, IDs, and hashes are retained. Of 300 displayed confidence values, 272 equal 1.0; these values are not interpreted as calibrated class probabilities.

The archived broad F1 is 0.863. Table~\ref{tab:llm_audit_exploratory} retains the hosted result beside Qwen3-8B solely for exploratory comparison. Both use the same single-coder audit. This table supplies neither a stable hosted-model benchmark nor evidence that a general-purpose endpoint is adequate for the full measurement task.

\begin{table}[H]
\centering
\caption{\textbf{Exploratory hosted-client comparison on the stratified 300-snippet audit}}
\label{tab:llm_audit_exploratory}
\begin{threeparttable}
\footnotesize
\begin{tabularx}{\textwidth}{@{}Xrr@{}}
\toprule
Audit label & Qwen3-8B F1 & Hosted Gemini F1 \\
\midrule
FX derivatives & 0.791 & 0.893 \\
Interest-rate derivatives & 0.826 & 0.927 \\
Commodity derivatives & 0.952 & 0.977 \\
Foreign-debt hedging & 0.148 & 0.424 \\
Broad hedging & 0.803 & 0.863 \\
Negation/non-use & 0.911 & 0.965 \\
Context-only disclosure & 0.548 & 0.811 \\
Hedge-accounting only & 0.408 & 0.841 \\
Unclear & 0.108 & 0.077 \\
\bottomrule
\end{tabularx}
\begin{tablenotes}
\footnotesize
\item The stratified audit supports diagnostic comparison rather than population prevalence estimation. Qwen3-8B used deterministic decoding and a recorded model revision. The Antigravity-hosted Gemini run is retained as a client-hosted benchmark because an exact API revision was unavailable.
\end{tablenotes}
\end{threeparttable}
\end{table}

The archived A100 experiment evaluates epochs 15, 20, 25, and 30. Mean weak-label F1 is 0.852234, 0.852417, 0.851299, and 0.850621. The absolute mean-F1 gain from epoch 15 to 20 is 0.000183, below the implemented 0.002 plateau threshold. Channel-selected checkpoints remain epochs 20, 20, 25, and 15. These are archived development choices. They are not independent tests of new hyperparameters or evidence of a prospectively optimized production system.

\end{document}